\PassOptionsToPackage{square,sort,comma,numbers}{natbib}

\documentclass{article}  

\usepackage[preprint]{nips_2018}

\usepackage{amsmath}
\usepackage{algorithm}
\usepackage[noend]{algpseudocode}
\usepackage{microtype}
\usepackage{balance}
\usepackage{graphicx}
\usepackage[sort]{cite}

\title{Learning Sharing Behaviors with Arbitrary Numbers of Agents}

\author{%
  Katherine Metcalf\\
  Indiana University\\
  Bloomington, IN \\
  \texttt{metcalka@indiana.edu} \\
  \And
  Barry-John Theobald \\
  Apple Inc.\\
  Cupertino, CA \\
  \texttt{barryjohn\_theobald@apple.com} \\
  \AND
  Nicholas Apostoloff \\
  Apple Inc.\\
  Cupertino, CA \\
  \texttt{napostoloff@apple.com}\\
}

\begin{document}

\maketitle

\begin{abstract}
We propose a method for modeling and learning turn-taking behaviors for accessing a shared resource.  We model the individual behavior for each agent in an interaction and then use a multi-agent fusion model to generate a summary over the expected actions of the group to render the model independent of the number of agents.  The individual behavior models are weighted finite state transducers (WFSTs) with weights dynamically updated during interactions, and the multi-agent fusion model is a logistic regression classifier.

We test our models in a multi-agent tower-building environment, where a Q-learning agent learns to interact with rule-based agents.  Our approach accurately models the underlying behavior patterns of the rule-based agents with accuracy ranging between 0.63 and 1.0 depending on the stochasticity of the other agent behaviors.  In addition we show using KL-divergence that the model accurately captures the distribution of next actions when interacting with both a single agent (KL-divergence $<$ 0.1) and with multiple agents (KL-divergence $<$ 0.37).  Finally, we demonstrate that our behavior model can be used by a Q-learning agent to take turns in an interactive turn-taking environment.
\end{abstract}

\vspace{2mm}

\section{Introduction}\label{sec:introduction}

Turn-taking plays an important role in interactions between people \citep{schegloff:2007:book,schegloff:2001:handbook}.  Conversational partners take turns at speaking, and someone wishing to take a turn might use body language to signal that they have something to say \citep{beattie1981interruption}.  Drivers take turns occupying segments of the road while avoiding collisions, and someone wishing to enter an occupied lane can use turn-signals to indicate their intention.  There has been a large amount of work covering turn-taking in conversational agents where interruptions are recoverable \citep{cafaro2016effects,raffensperger:2012:metric}, but our focus is on applications that require strict turn-taking behavior, in which collisions can be catastrophic.

There are expected turn-taking patterns that are defined by social-cultural norms and explicit rules.  Adhering to social-cultural norms can significantly enhance the performance of agents acting within a group \citep{epstein:2001:ce}.  Deviations from these patterns may require turn-taking to be re-negotiated, which slows down task completion and can result in adverse outcomes. The personalities and individual attributes of those participating in an interaction will influence the extent to which norms are followed. Norm and personality-based attributes are hard to define formally, are complex, and can involve contradictory requirements and constraints \citep{delgado:2002:ai}. Therefore, the effects of social-cultural and personality-based attributes are best discovered through observation and experience.

We treat multi-agent interactions as a three-step process: 1) select \textbf{what} to do (e.g. change lane), 2) identify \textbf{when} to do it (e.g. merge in front of or behind the occupying vehicle), and 3) determine \textbf{how} to do it (e.g. execute the path).  Decomposing multi-agent interactions into a process of identifying \textbf{what}, \textbf{when} and \textbf{how} results in a hierarchical model with components that can be tackled in semi-isolation; the more general task-agnostic \textbf{when} component, which is most influenced by social-cultural norms, is decoupled from the more task-specific action selection and execution components.  This decomposition allows the regularity imposed by social-cultural norms and turn-taking patterns to be modeled in a way that is applicable across interactive domains in which agents should not take turns at the same time.

Our focus is on modeling the turn-taking behavior of agents with different behavioral \emph{types} (i.e.\ erratic, passive, and aggressive)  such that a learning-agent can learn to collaboratively build towers.  Our behavior model is able to accurately predict the type of turn-taking action other agents will take, and a Q-learning agent can use our behavioral model to learn a turn-taking policy \citep{raffensperger:2012:rewards} that can be used in interactions involving varying numbers of agents.

The rest of this paper is organized as follows:  In Section \ref{sec:related-work} we present the relevant background.  Section \ref{sec:domain} presents an overview of the technologies and learning algorithms used.  Section \ref{sec:experiments} describes the experiments conducted to validate our models.  The results are described in Section \ref{sec:results}, and Section \ref{sec:conclusions} provides a summary of our work and suggestions for future work.

\begin{figure*}
\centering
\begin{tabular}{ccc}
\includegraphics[width=0.2\paperwidth]{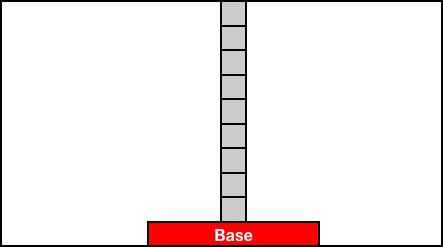} &
\includegraphics[width=0.2\paperwidth]{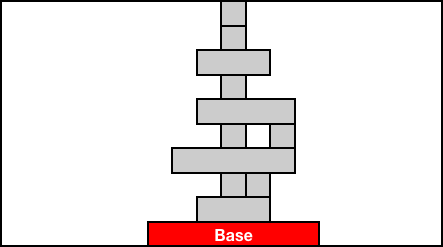} &
\includegraphics[width=0.2\paperwidth]{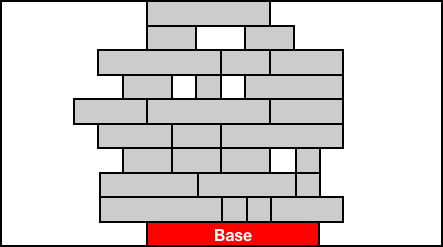} \\
\end{tabular}
\caption{The tower building domain with increasingly complex towers scoring nine (left), 22 (center), and 73 (right).  In our experiments the width and height of the world are 100 bricks (not shown here), and the base is of size eight bricks.}
\label{fig:tower-examples}
\end{figure*}

\section{Related Work} \label{sec:related-work}

Prior work on learning acceptable social-cultural norms typically focuses on evaluating how long it takes a group of artificial agents to converge upon a set of norms \citep{vouros:2017:aamas}, and techniques fall into one of several categories: imitation, normative advice, machine learning, and data-mining \citep{savarimuthu:2011:info}. Such social learning models typically assume that the agents involved are able to communicate any required information for learning a policy \citep{vouros:2017:aamas}, an assumption that may not apply to all domains. We assume that there is a \emph{pre-existing set of norms that the agent must learn}, so there is no process of norm negotiation, and these norms must \emph{be learned without access to the strategies of the other agents}. For example, it is not feasible to exit a moving car to ask another driver how they intend to perform a maneuver.  Therefore, we take the approach of learning behavioral styles by imitating the behavior of other agents for the purpose of predicting future actions. Following \citep{mukherjee:2008:aamas,sen:2007:ijcai,vouros:2017:aamas,yu:2013:aamas}, we use Q-learning algorithms to learn norms.

Our work has characteristics in common with Sequential Social Dilemmas (SSD): the interaction is temporally extended, decisions to cooperate or defect occur quasi-simultaneously, and decisions must be made given only partial information about the activities of the other players \citep{leibo:2017:aamas}. However, such approaches do not explicitly account for the influence of social-cultural norms or the personalities of the other agents participating. Instead the focus, as it is for Matrix Game Social Dilemmas (MGSD), is on how to select action strategies that are most socially beneficial.  In our domain, the best strategy is \emph{cooperating with turn-taking norms by alternating access to a resource}. Our goal is to learn how to cooperate. Therefore, we do not take a game theoretic approach and instead focus on modeling when others are likely to take a turn. 

Our model maintains beliefs of a set of hypothetical behaviors that may be observed in other agents. Maintaining a set of beliefs over possible next actions is one strategy for handling situations in which there is no opportunity for coordination prior to entering the task domain \citep{albercht:2016:ai,albrecht:2014:uncertainty,albrecht:2017:aamas,barrett:2015:aaai,barrett:2011:aamas,chandrasekaran:2014:aamas,southey:2005:uncertainty}. Typically, behaviors are specified as a set of types (i.e. blackbox mappings from interaction histories to probability distributions). When the types sufficiently represent the true underlying behaviors, they can be used to rapidly adapt strategies for effective interactions \citep{barrett:2015:aaai}. We handle the issue of non-parameterized types, as identified in \citep{albrecht:2017:aamas}, by designing models that can be refined to reflect the specific behavior types being observed.

Melo and Sardinha \citep{melo:2016:aamas} propose an approach for managing ad-hoc teamwork where teammates are treated as a single meta-agent, and actions are predicted for the meta-agent at each time step. The learning-agent then selects a joint action that includes its own action and the predicted action for its teammate. We take a similar approach when selecting a next action for our learning-agent --- the Q-function considers the expected actions at the group level, and we include an estimate of what our learning-agent will do when making predictions about what other agents will do.

\section{Learning Sharing Behaviors} \label{sec:domain}

In this work, we are interested in modeling \textbf{when} agents can and will take a turn using a limited capacity shared resource.  The complexity of requiring agents to select a goal (i.e.\ \textbf{what} to do) is removed by choosing a problem domain that has only a single goal.  This allows us to focus specifically on \textbf{when} agents should take a turn in an interaction.

We evaluate our turn-taking behavior models with a tower building game (see Figure \ref{fig:tower-examples}), where the objective is to build complex towers (measured as the number of \emph{bricks} used to build the tower) by placing \emph{blocks} (formed of $1 \times m$ bricks, with $1 \leq m \leq 5$). The game always begins with a fixed-size base in the center of the world.  The game is complete when all bricks have been used, the tower reaches a pre-defined maximum height, or the tower collapses.  The tower collapses when more than one agent places a block at a time (even if the block placements do not collide) or when a block placement offsets the center of gravity of the tower so that it is no longer supported by the structure.

There are three \emph{types} of rule-based agents with varying degrees of stochasticity that build towers in this domain.  Specifically, the agent \emph{types} are a simple deterministic agent that plays passively, a semi-stochastic agent that plays aggressively, and a fully-stochastic agent that plays erratically.  These \emph{types} encompass both different styles of turn-taking and varying degrees of difficulty for predicting next actions.  In this work, the agent \emph{types} and the \emph{type} assignment for each agent are known \emph{a priori},  but in a more general approach we could cluster agents based on observed behavior to define types.  In addition, we could represent agent behavior \emph{type} as a distribution over \emph{types} (see Section \ref{sec:conclusions}). 

The rule-based agents are formed of:
\begin{itemize}
\item{A shared tower building policy learned using a deep Q-network (DQN) that defines \textbf{how} to place blocks to build towers.}
\item{A distribution over block sizes that is unique to each agent and that the agent draws from to play the game.}
\item{A unique behavior policy that defines the agent \emph{type} (Table \ref{tab:next-actions}).  This policy is modeled as a distribution over next actions, where possible actions are: 1) do nothing --- pass ($p$), 2) indicate an intention to place a block ($i$), or 3) access the resource --- place a block on a tower ($a$). }
\end{itemize}

At each time step in the simulation, all agents choose a type of turn to take:  the rule-based agents by sampling from their assigned policy and the learning-agent from its learned policy.  If an agent chooses to place a block, the shared tower building policy is used to determine where to place the block on the tower.

Agents must strictly take turns to avoid collapsing the tower.  For this reason, agents may not place a block in their first turn, rather they must either pass or indicate their intention to place a block. This restriction avoids over-penalizing the learning-agent because, without context, it can only guess the first move of each agent.

\begin{table*}[h!]
\centering
\tiny
\begin{tabular}{ |l|c|c|c||c|c|c||c|c|c| }
\hline
& \multicolumn{9}{|c|}{Next Action}\\\hline
 & \multicolumn{3}{c||}{Passive} & \multicolumn{3}{|c||}{Aggressive} & \multicolumn{3}{|c|}{Stochastic}\\
 \hline
Previous Action & p & i & a & p & i & a & p & i & a\\
\hline
Only agent $a_{i}$ indicated & 0.0 & 0.0 & 1.0 & 0.0 & 0.0 & 1.0 & 0.05--0.15 & 0.05--0.15 & 0.75--0.85 \\
Agent $a_{i}$ and other agents indicated & 0.0--0.05 & 0.95--1.0 & 0.0 & 0.0 & 0.0 & 1.0 & 0.55--0.65 & 0.35--0.45 & 0.0--0.05\\
Pass & 0.0--0.05 & 0.95--1.0 & 0.0 & 0.0--0.05 & 0.95--1.0 & 0.0 & 0.35--0.45 & 0.50--0.60 & 0.05--0.10\\
Place block & 1.0 & 0.0 & 0.0 & 0.05--0.15 & 0.85--0.95 & 0.0 & 0.05--0.15 & 0.60--0.70 & 0.20--0.30\\
\hline
\end{tabular}
\caption{The range of values used to generate the distribution of next actions for rule-based agents.  To generate an agent, select the desired behavior style, sample values in the ranges shown and normalize to form a distribution over next actions.}
\label{tab:next-actions}
\end{table*}

\begin{figure*}
\centering
\includegraphics[scale=0.07]{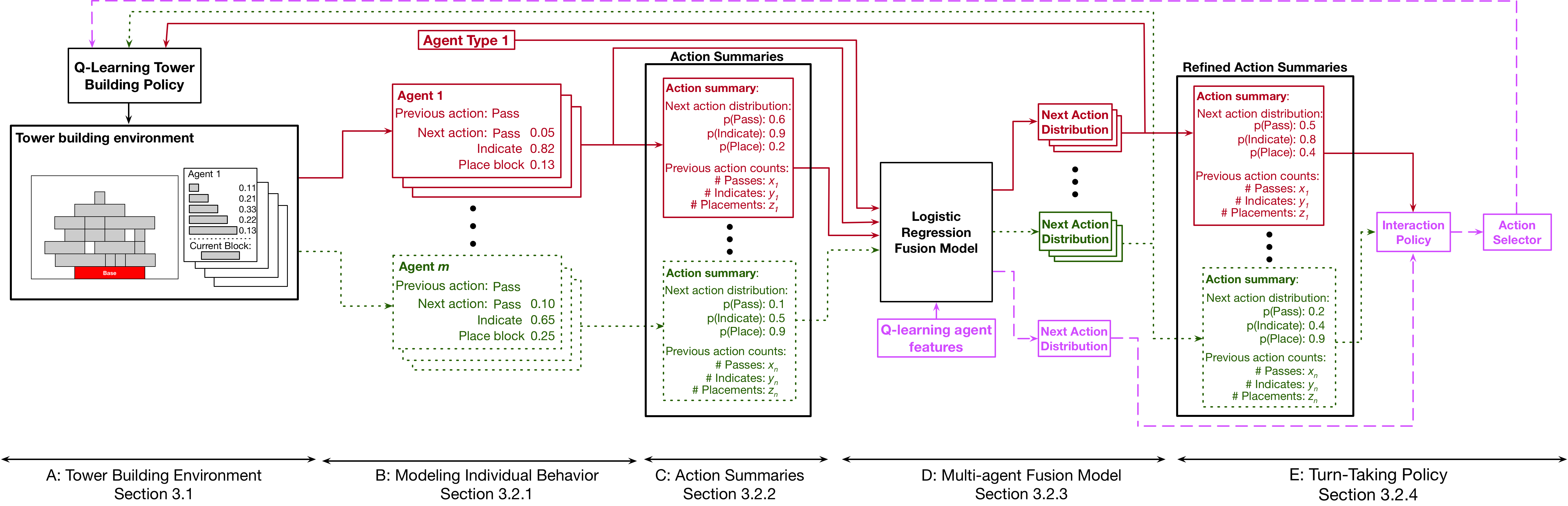}
\caption{An overview of our model for interactive behavior modeling.  Components that relate to modeled agents of different behavior \emph{types} are colored using red (solid) and green (dotted) lines, and the learning-agent is colored purple (dashed).  The example shown is for an agent of \emph{Type 1}; the process repeated for all agents of all \emph{types}.}
\label{fig:system-overview}
\end{figure*}

\subsection{Learning to Build Towers}\label{sec:tower-builder}

A tower building policy is learned offline using the DQN architecture described in \citep{mnih:2013:dqn}, which has been used previously to  learn to play games successfully.  The purpose of this policy is to learn where to place blocks to build towers.  Once the tower building policy is learned, it is used by all agents, but each agent is assigned a different distribution of block sizes to draw from to provide a degree of uniqueness among the agents.

The tower building policy takes as input the block size for the current turn and a binary image that encodes where bricks have been placed.  No feedback is provided about the quality of block placements during an episode, rather the reward is applied only after the episode is complete.  The reward is equal to the number of bricks in the tower, which encourages building both upwards and outwards to construct more complex towers. See Figure \ref{fig:tower-examples} for examples of towers of varying complexity.

\subsection{Modeling Interactive Behaviors}

An overview of our behavior model in the tower building domain is shown in Figure \ref{fig:system-overview}.  The purpose of the behavioral model is to represent the observed behavior patterns of the agents in a group such that a Q-learning agent can take turns without causing collisions.  Broadly, the behavior model performs the following:
\begin{enumerate}
\item{For each rule-based agent in isolation, predict a distribution over its next action based only on its previous action (Figure \ref{fig:system-overview} B and Section \ref{sec:individual-behavior}).}
\item{Summarize the predictions by agent \emph{type} (Figure \ref{fig:system-overview} C and Section \ref{sec:action-summary})}
\item{Refine the individual action predictions from step 1 using the summaries from step 2 as input to a multi-agent fusion model (Figure \ref{fig:system-overview} D and Section \ref{sec:fusion-model}).}
\item{Summarize the refined action predictions from step 3 and input to the interaction policy to select a next action (Figure \ref{fig:system-overview} E and Section \ref{sec:rl-policy}).}
\item{Place a block using the Q-learning tower building policy if the selected action is to place a block (Figure \ref{fig:system-overview} A and Section \ref{sec:tower-builder}).}
\item{Roll-out the behavior model for $n$ time-steps to maximize the expected future reward by using the most likely action at time $t$ as the previous action for time $t+1$ for all agents.}
\end{enumerate}

\subsubsection{Modeling Individual Behavior} \label{sec:individual-behavior}

\begin{figure}
\centering
\includegraphics[scale=0.22]{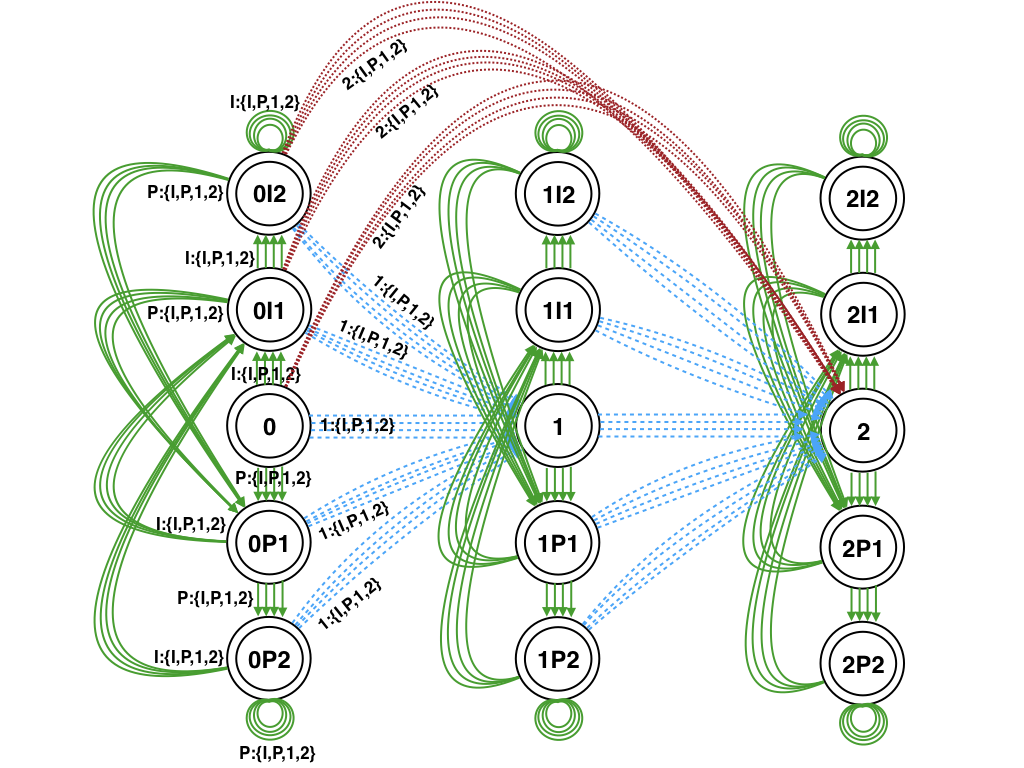}
\caption{The behavior model for two steps in our tower building domain.  Each column of states represents a certain number of bricks placed by the agent.  The colors on the arcs signify that no block was placed in the turn (solid green), a block of size one was placed (dashed blue), or a block of size two was placed (dotted red).  The state labels reflect the actions, e.g.\ ``mI2'' signifies that after placing $m$ bricks on the tower, the agent has indicated at least twice in a row.  The arc labels denote input:output pairs, where the input symbol represents an observed action and the corresponding output symbol represents the predicted next action.  With this structure we assume that the probability of indicating or passing two or more times is the same.  This can be altered by growing the WFST vertically as required.  For brevity, the output arc labels are grouped into sets, and the weights are not included.}
\label{fig:wfst-full}
\end{figure}

The individual agent behavior models are weighted finite state transducers (WFSTs).  A WFST is a network of states connected by directed arcs that each map an input symbol to an output symbol with an associated cost. The structure of the WFSTs used in this work is depicted in Figure \ref{fig:wfst-full}, which shows the possible state sequences for an agent that has been allocated two bricks. This model can be expanded to any arbitrary number of bricks by adding columns of states and the appropriate arcs between the states.  In our domain, a state represents the number of bricks an agent has placed and the action that it took in the last turn.  In Figure \ref{fig:wfst-full}, the input symbol on each arc represents an \emph{observed action}, and the corresponding output symbol represents the \emph{predicted next action}. Each state has multiple output arcs (one for each possible input-output action pair), therefore the set of arcs to each next state reflects the distribution over next actions given a particular previous action that was observed.  WFSTs allow us to model agent behavior as a function of the progression of a game. For example, as a tower nears completion, an eager agent might become more cautious to avoid losing a large reward, so the distribution over next actions is dependent on where in an episode the agent currently is.  

In the context of modeling behavior, the sequence of actions taken by an agent is a path through the WFST.  The likelihoods on the arc transitions are learned using an approach inspired by \emph{learning to search}, which is used to simplify structured joint prediction tasks in computational linguistics \citep{ross2011reduction}.  We use the exponential moving average to update the weights as follows:
\begin{equation}
w_{t} = \left\{\begin{array}{ll}
c_{t} & \mathrm{if\ }w_{t} \mathrm{\ is\ uninitialized}\\
\eta c_{t} + \left(1 - \eta\right)w_{t-1} & \mathrm{otherwise},
\end{array}\right.
\label{eqn:exp_average}
\end{equation}
where $w_t$ is the weight at time $t$, and $c_t$ is the observed cost at time $t$.  In any given state, there are three arcs that correspond to a particular observed action (one arc for each next action).  The cost $c_t$ is 0 for the arc corresponding to the observed next action and 1 for the remaining two arcs.  The weight, $w_t$, is uninitialized the first time a transition between a particular pair of states occurs.  The update in Equation \ref{eqn:exp_average} is used because it provides a degree of robustness to noisy observations as the influence of older versus more recent observations can be tuned by changing $\eta$ as needed (we used $\eta = 0.1$), and the arc weights can be updated without having to track the entire history associated with that arc.

\subsubsection{Action Summaries} \label{sec:action-summary}

To render the interaction policy independent of the specific number of agents in the interaction, we summarize the predictions for the next action distributions by agent \emph{type}.  The summary for each agent \emph{type} is formed of two parts: 1) the probability of observing any agent of that \emph{type} taking each of the available actions in the next turn, and 2) a histogram of the actions taken by the agents of that \emph{type} in the previous turn.  The entries in the summary for any agent \emph{types} that are not present in the interaction are set to zero.  Since we are summarizing the actions by agent \emph{type}, we are dependent on only the number of actions and the number of behavior \emph{types}, not the number of agents.

\subsubsection{Multi-agent Fusion Model} \label{sec:fusion-model}

The behavior model maintains an agent-specific WFST to model the behavior of each (rule-based) agent it is participating with.  While the predictions produced by a WFST are based only on the actions that a particular agent has taken, interacting agents will influence one another.  Thus, the action predictions are refined using a multi-agent fusion model that takes into account the predicted next actions and the behavior \emph{type} of all agents. 

The multi-agent fusion model uses multinomial logistic regression to predict a refined distribution over next actions for each agent for the next time step. The model parameters are learned using the cross-entropy loss function and stochastic gradient descent.  For each agent, the model takes as input a flag representing the agent's type, a one-hot vector encoding the agent's last action, the WFST predictions for that agent, and the \emph{action summaries} for each agent \emph{type}.  The output is a refined distribution over next actions for the given agent, refined with respect to the behaviors of all other agents.  Once the predicted action distribution has been refined for all agents, the action likelihoods are then re-summarized by agent \emph{type}, and this summary along with the action distribution for the learning-agent are input to the interaction policy to select an action for the learning-agent.

The dimensionality of the feature space consumed by the fusion model (i.e.\ $n_{a}(2 \times n_{t} + 2) + 1$, where $n_{t}$ is the number of agent types, and $n_{a}$ is the number of actions) grows linearly with the number of actions and agent \emph{types}, but this is small compared to any approach attempting to represent the domain as an image (e.g.\ to obtain a fixed sized feature vector for neural network training).  The low dimensionality greatly reduces the amount of data we need to learn models of turn-taking behaviors and is what allows us to use a simple model like multinomial logistic regression.

\subsubsection{Turn-Taking Policy} \label{sec:rl-policy}

We used a Q-learning algorithm to learn a policy for choosing which action $\{p, i, a\}$ a learning-based agent should take when collaborating with other rule-based agents. The state-space for the policy is continuous, but relatively simple and all values fall on the range  [0, 1], which represent the probability of each action type being observed in the next time step.  The simplicity of the state-space allows us to predict the discounted expected reward for each given action using linear function approximation (Figure 6.7, pg.\ 137) in \citep{sutton:1998:book}, where the inputs are the (rolled-out) refined action summaries for each rule-based agent \emph{type} and the (rolled-out) refined distributions over next actions for the learning-agent, and output is the Q-value representing the quality of each action.   At the end of each game, the agent receives a reward equivalent to the number of successful block placing actions it had taken. A block placement was considered successful if the Q-learning agent did not attempt to place a block at the same time as a rule-based agent. If the Q-learning agent attempted to place a block at the same time as a rule-based agent then the total score was decreased by one. This means that to achieve the maximum possible reward during each game, the Q-learning agent must successfully place all allocated blocks (i.e.\ learn \textbf{when} to take turns).

The turn-taking policy only handles the selection of the action $\{p, i, a\}$.  If the learning-agent selects to access the resource (i.e.\ chooses action $a$) it then uses the tower-building policy, discussed in Section \ref{sec:tower-builder}, to place the block on the tower (i.e.\ \textbf{how} to take the turn).  The tower building policy used by the learning-agent is the same policy used by the rule-based agents.

\section{Experiments} \label{sec:experiments}

There are three behavioral components of our learning-agent that require training: 1) the general WFSTs for each behavior \emph{type} that are used as the basis of the behavior models for individual agents, 2) the multi-agent fusion model that consumes and refines action summaries, and 3) the interaction policy that consumes the refined action summaries and selects the next action for the learning-agent.  Each subsequent step in the behavior model relies on the previous stages and so we adopt a sequential approach to training.  This is outlined in the following sections.

\subsection{Training general behavior models}\label{sec:behavior-model-training}

The learning-agent must first learn a prototypical behavior model for each behavior \emph{type}.   To do this the learning-agent creates a set of uninitialized WFSTs (Figure \ref{fig:wfst-full}), where the cost on all arcs is 1.  Next it observes a series of rule-based agents interacting, where there are 30 games of each permutation of one, two and three agent \emph{type} combinations (210 interactions in total).  Each rule-based agent is created by first selecting the desired behavior \emph{type}, then sampling the values for the next action likelihoods from the ranges given in Table \ref{tab:next-actions}, and finally normalizing to form a distribution over next actions.  During the interactions, the learning-agent is only observing the game-play, it is not interacting with the other agents.  The aim is only to update the weights on the WFST arcs representing the specific agent \emph{types} as turns are taken by the rule-based agents.

After training the prototypical WFST for each agent \emph{type}, these models then form the basis of the behavior models used by the learning-agent to represent the behavior of new rule-based agents.  A new rule-based agent is assigned the prototypical model corresponding to its \emph{type}, and the weights on the arcs are adapted from the initial `prototypical' values to the values that represent the specific agent behavior as the learning-agent interacts with the rule-based agent(s).  Note, the rule-based agents have behavior \emph{types} known \emph{a priori}, which means that we do not need to estimate agent \emph{type} from behavior and we can control and manipulate precisely the turn-taking behaviors of these agents.  

\subsection{Training the multi-agent fusion model}\label{sec:fusion-model-training}

The behavior model contains a single multi-agent fusion model that is trained while observing rule-based agents interacting with one another.  Again, during this observation phase, the learning-agent is not interacting with the rule-based agents.  Each rule-based agent is assigned the general behavior model corresponding to its \emph{type}, and these are adapted during the interactions to match the distribution over next actions for each particular agent.  The action summaries from the next action predictions (Section \ref{sec:action-summary}) are used as the training inputs for the multinomial logistic regression, and the observed next actions are the ground-truth output targets during training.  As in Section \ref{sec:behavior-model-training}, the learning-agent observes each permutation of one, two and three agent \emph{types} interact for 30 games.

We train a single multi-agent fusion model that is used in all subsequent interactions, regardless of the number of interacting agents.  Unlike the WFSTs, the multi-agent fusion model is not updated or refined as it observes agents interact\footnote{There exist online learning algorithms for multinomial logistic regression that would allow the model to update itself based on current observations and adjust to any changes in turn-taking patterns if we ever decide we need that: \texttt{https://lingpipe.files.wordpress.com/2008/04/lazysgdregression.pdf}}.  The multi-agent fusion model is used to represent and account for the more stable turn-taking patterns that stem from social-cultural norms while the single-agent behavior models are used to account for individual differences. 

\subsection{Training the interaction policy}

After learning the general behavior models and the multi-agent fusion model, the interaction policy is trained by having the learning-agent interact with various combinations of rule-based agents.  During these interactions, we also evaluate how well the weights on the arcs reflect the next action distributions of the rule-based agents (see Table \ref{tab:behaviorresults}), and how well the multi-agent fusion model refines these individual next action predictions (see Table \ref{tab:fusionresults}).  We measure performance in terms of accuracy of predicting the next action for the rule-based agents, and using the KL divergence the extent to which the approximated action distributions differ from the true action distributions.  Note, the results in Tables \ref{tab:behaviorresults} and \ref{tab:fusionresults} are produced while the individual behavior models are being adapted --- we would expect higher mean accuracy and lower mean KL divergence if evaluating only after the models had converged.

Whilst training the interaction policy, the exploration rate of the Q-learner, $\epsilon$, was initially set to $1.0$ for the first 50 games, then was decayed linearly to $0.0001$ over 400 games, and finally remained constant for the remaining games. The learning rate $\alpha$ was decayed according to the same schedule.  We measure how well a Q-learning agent can use our behavior model to make decisions about \textbf{how} to take actions in an interactive task by reporting the mean return, averaged over 50 trials for each agent \emph{type} combination (see Figures \ref{fig:PassiveAgentLearningCurve} -- \ref{fig:StochasticAgentLearningCurve}, and Figures \ref{fig:AllAgentsLearningCurve} and \ref{fig:MultipleAgentsSameType}.

We compared our behavioral model as the input to the Q-learning agent against a baseline state space. The baseline state space is fixed relative to the number of agents participating in the interaction, and the dimensionality is equal to the number of agents participating in the interaction (including the Q-learning agent). The state space represents the turn-taking actions observed during the last time step. For example, if the Q-learning agent is playing against one other agent, and in the last turn the Q-learning agent indicated and the other agent placed a block, then the state space is  $[i, a]$. We chose this state representation as our baseline because it is the same state representation used by the rule-based agent, so the Q-learning agent should be able to learn appropriate behavioral responses given the same state representation, and this baseline serves as a point of comparison for our new model.  Unlike our behavior model that is independent of the number of agents in the interaction, this baseline state space has prior knowledge of the number of agents.  Also, the learned policy using the baseline state space is only applicable for this specific number of agents.

\section{Results} \label{sec:results}

\begin{table}
\centering
\begin{tabular}{ |l|l|l|l|l|l| }
\hline
\multicolumn{6}{|c|}{Single Agent} \\
\hline
\multicolumn{2}{|c|}{Passive} & \multicolumn{2}{|c|}{Aggressive} &  \multicolumn{2}{|c|}{Stochastic}\\
\hline
Acc.&KLD&Acc.&KLD&Acc.&KLD\\
\hline
1.00&0.0001&0.900&0.064&0.778&0.077\\
\hline
\multicolumn{6}{|c|}{Pairwise Agents} \\
\hline
\multicolumn{2}{|c|}{Passive} & \multicolumn{2}{|c|}{Aggressive} &  \multicolumn{2}{|c|}{Stochastic}\\
\hline
Acc.&KLD&Acc.&KLD&Acc.&KLD\\
\hline
0.907&0.011&0.791&0.213&0.7&0.237\\
\hline
\multicolumn{6}{|c|}{All Agents} \\
\hline
\multicolumn{2}{|c|}{Passive} & \multicolumn{2}{|c|}{Aggressive} &  \multicolumn{2}{|c|}{Stochastic}\\
\hline
Acc.&KLD&Acc.&KLD&Acc.&KLD\\
\hline
0.890&0.021&0.693&0.003&0.63&0.015\\
\hline
\end{tabular}
\caption{Mean accuracy (acc.) and KL Divergence (KLD) for the next action predictions from single agent WFST models. All scores are averaged across 50 independent runs for all combinations of agent types.}
\label{tab:behaviorresults}
\end{table}

\begin{table}
\centering
\begin{tabular}{ |l|l|l|l|l|l| }
\hline
\multicolumn{6}{|c|}{Single Agent} \\
\hline
\multicolumn{2}{|c|}{Passive} & \multicolumn{2}{|c|}{Aggressive} &  \multicolumn{2}{|c|}{Stochastic}\\
\hline
Acc.&KLD&Acc.&KLD&Acc.&KLD\\
\hline
1.0&0.05&0.95&0.37&0.8&0.2\\
\hline
\multicolumn{6}{|c|}{Pairwise Agents} \\
\hline
\multicolumn{2}{|c|}{Passive} & \multicolumn{2}{|c|}{Aggressive} &  \multicolumn{2}{|c|}{Stochastic}\\
\hline
Acc.&KLD&Acc.&KLD&Acc.&KLD\\
\hline
0.931&0.003&0.868&0.005&0.735&0.015\\
\hline
\multicolumn{6}{|c|}{All Agents} \\
\hline
\multicolumn{2}{|c|}{Passive} & \multicolumn{2}{|c|}{Aggressive} &  \multicolumn{2}{|c|}{Stochastic}\\
\hline
Acc.&KLD&Acc.&KLD&Acc.&KLD\\
\hline
0.907&0.012&0.693&0.003&0.63&0.024\\
\hline
\end{tabular}
\caption{Mean accuracy (acc.) and KL Divergence (KLD) for the refined next action predictions produced by multi-agent fusion models. All scores are averaged across 50 independent runs for all combinations of agent types.}
\label{tab:fusionresults}
\end{table}

We first consider the performance of the learning-agent interacting with a single rule-based agent.  This allows us to determine an upper bound on expected accuracy of the next action prediction given the different agent \emph{types} without confounding effects that arise when interacting with multiple rule-based agents.  As one would expect, the accuracy with which the next action can be predicted varies as a function of the stochasticity of the agent being observed.  The behavior of the deterministic agent is learned perfectly, and the behavior model does well at modeling the distribution of the next actions for the semi-stochastic and fully-stochastic agents (accuracy $>0.78$ and KLD $< 0.08$, see Table \ref{tab:behaviorresults}).  Also, the fusion model used to refine the predictions improves the performance as expected (accuracy $>0.8$ and KLD $< 0.2$, see Table \ref{tab:fusionresults}).

\begin{figure}
\centering
\includegraphics[scale=0.24]{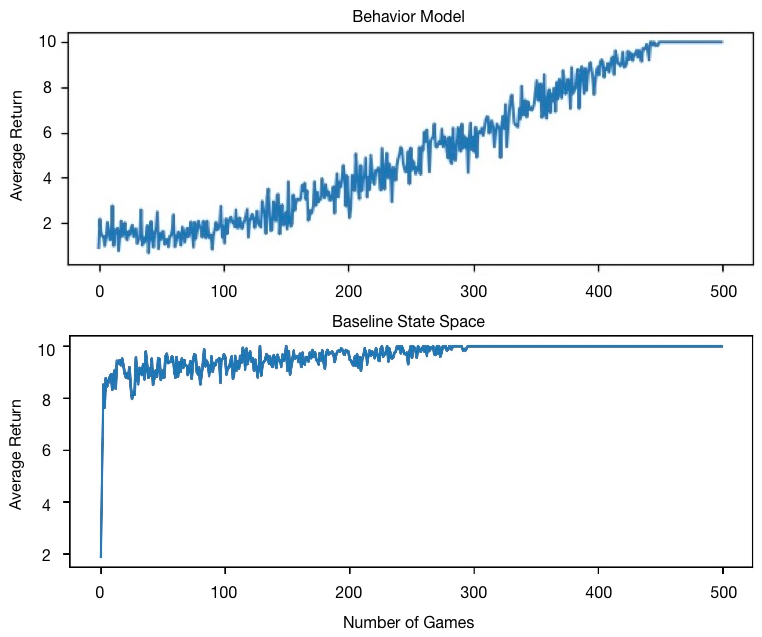}
\caption{Learning curves for the Q-learning agent playing with the passive rule-based agent with $\alpha=0.01$ and $\gamma=0.90$.  The agent is able to learn a perfect policy using both state spaces, but the rate of learning is faster for the baseline.}
\label{fig:PassiveAgentLearningCurve}
\end{figure}

\begin{figure}
\centering
\includegraphics[scale=0.24]{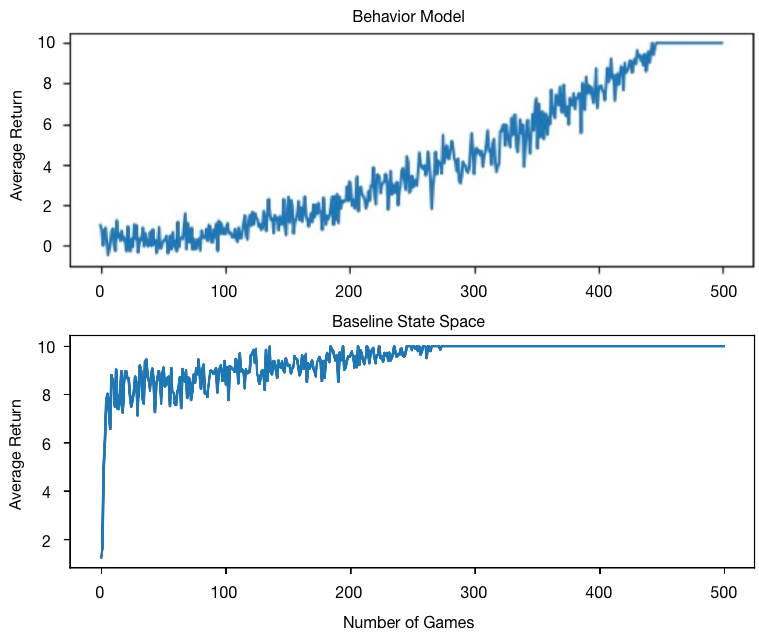}
\caption{Learning curves for the Q-learning agent playing with the aggressive rule-based agent with $\alpha=0.01$ and $\gamma=0.99$. The agent is able to learn a perfect policy using both state spaces, but the rate of learning is faster for the baseline.}
\label{fig:AggressiveAgentLearningCurve}
\end{figure}

When using the behavior predictions as input to the Q-learning agent for decision making, an optimal policy is learned for the passive and aggressive rule-based agents for both the baseline state representation and the behavior model state representation (Figures \ref{fig:PassiveAgentLearningCurve} and \ref{fig:AggressiveAgentLearningCurve}).  Despite differences in optimal gamma values (determined using a parameter sweep), the learning curves show similar trends across both figures. However, convergence to the optimal policy occurs more slowly for the behavioral model compared with the baseline state space, which is the result of the more complex structure of the state space returned by the behavior model.

\begin{figure}
\centering
\includegraphics[scale=0.24]{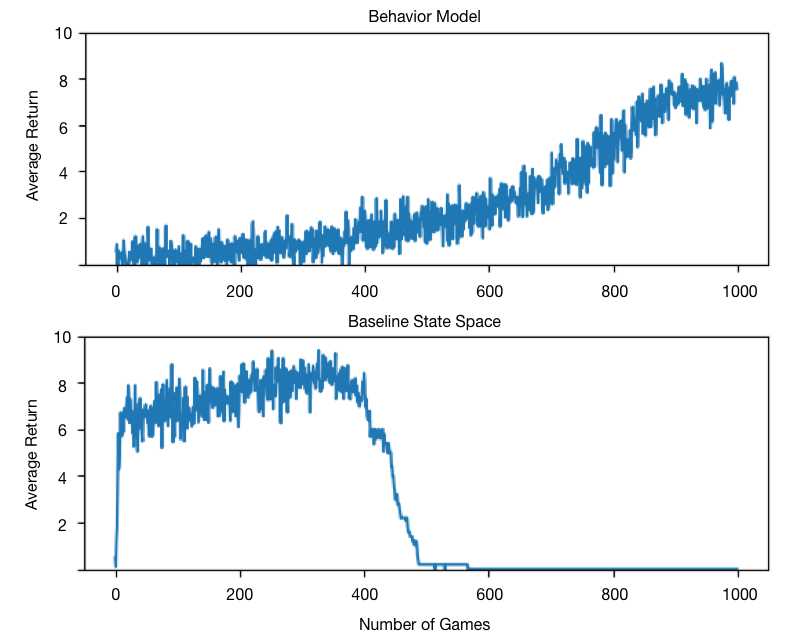}
\caption{Learning curves for the Q-learning agent playing with the stochastic agent with $\alpha=0.01$ and $\gamma=0.90$.  A policy is learned using the behavior model despite the erratic nature of the rule-based agent, but the agent fails to learn a policy with the baseline state space.}
\label{fig:StochasticAgentLearningCurve}
\end{figure}

\subsection{Interacting with a Single Rule-Based Agent}

\begin{figure}
\centering
\includegraphics[scale=0.42]{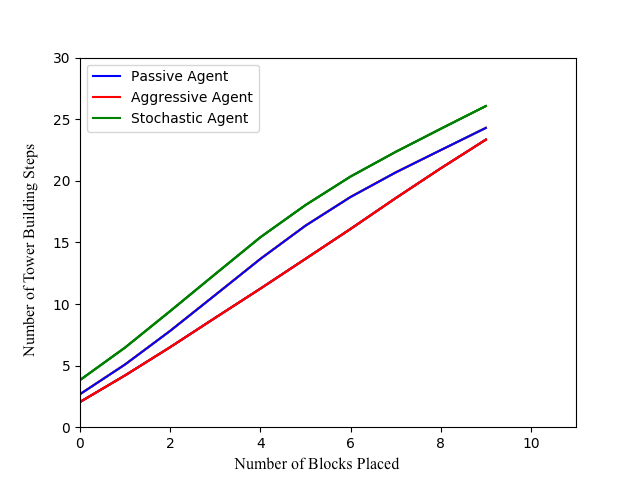}
\caption{The mean number of steps into the game before the learning-agent placed the $i^{th}$ block when playing agents of different behavior styles.  The near-linear curves show that the agents are not waiting until all other agents have placed all of their blocks before taking a turn.}
\label{fig:block_placment_steps}
\end{figure}

For the stochastic rule-based agent (Figure \ref{fig:StochasticAgentLearningCurve}), neither the baseline nor the behavioral model state representations learn an optimal policy, and it took longer to learn a policy that  converged for this agent than either the passive or the aggressive agent (approximately 1000 steps vs.\ 500 steps).  Interestingly, once $\epsilon$ decays to $0.0001$ (where the Q-learning agent rarely explores and relies heavily on its policy) performance drops for the baseline state representation due to the stochastic behavior of the agent, but the behavioral model continues to learn as the weights on the WFST arcs converge to the distributions defining the behavior.

The performance of the Q-learning agents when playing the stochastic rule-based agent indicates a difference in the usefulness of the baseline and the behavioral model representations. The decrease in performance using the baseline state representation after $\epsilon$ finished decaying suggests that some degree of randomness during action selection is beneficial when facing a highly random agent. Indeed, it appears that the behavior model state representation can account for the randomness of a single stochastic rule-based agent. Representing the state of the rule-based agent as a set of probabilities directly captures the uncertainty about the rule-based agent's next actions.

To ensure that the Q-learning agent was not learning to wait until the other agents had used all of their allocated blocks before placing its own blocks, we evaluated the average number of turns that were taken when the agent placed its $i^{th}$ block (Figure \ref{fig:block_placment_steps}).  The near-linear relationship between the number of game steps and the number of blocks placed by the Q-learning agent indicates that turn-taking behaviors are being learned.

\subsection{Interacting with Multiple Rule-Based Agents}

\subsubsection{Multiple Agents of Different Types}

\begin{figure}
\centering
\includegraphics[scale=0.24]{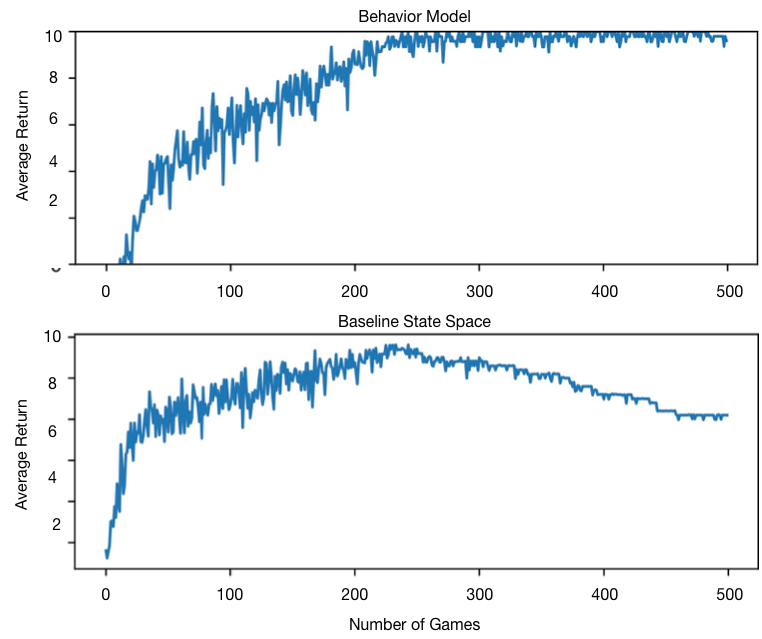}
\caption{Learning curves for the Q-learning agent playing with an agent of each behavior type with $\alpha=0.001$ and $\gamma=0.98$.  Notice that the agent using the behavior model learns an optimal policy despite the inclusion of the stochastic agent because the behavior of the other agents somewhat regulates the erratic behavior of the stochastic agent.}
\label{fig:AllAgentsLearningCurve}
\end{figure}

Learning behavior with multiple rule-based agents interacting with one another is more challenging, but the behavior model is able to capture the distribution over next actions when playing with each pair of agent types (accuracy $\geq 0.7$ and KLD $\leq 0.24$, see middle section of Tables \ref{tab:behaviorresults} and \ref{tab:fusionresults}) and when playing with all three agent types (accuracy $\geq 0.63$ and KLD $\leq 0.24$, see bottom section of Tables \ref{tab:behaviorresults} and \ref{tab:fusionresults}). In general, as the number of agents participating in the interaction increases, the performance of both the single agent WFSTs and the multi-agent fusion model decreases. However, the accuracy remains usefully high and the KL Divergence remains usefully low given the number of action combinations the rule-based agents can exhibit, as shown by the performance of the Q-learning agent in the top of Figure \ref{fig:AllAgentsLearningCurve}.  An interesting observation is that neither model converged to an optimal policy when playing only the stochastic agent (Figure \ref{fig:StochasticAgentLearningCurve}), but when a stochastic agent is playing with agents of other types, the behavior of the other agents regulates the erratic behavior of the stochastic agent and our behavior model does learn an optimal policy. However, the performance of the Q-learning agent using the baseline state representation does not converge to an optimal policy and decreases after $\epsilon$ reaches $0.0001$.

\subsubsection{Multiple Agents of the Same Type}

The previous experiments demonstrate that the behavior model is able to account for multiple agents, but where only a single agent of each behavior type is present.  Here we consider the result of the summarizing step when there are multiple agents of the same type present.

\begin{figure}
\centering
\includegraphics[scale=0.24]{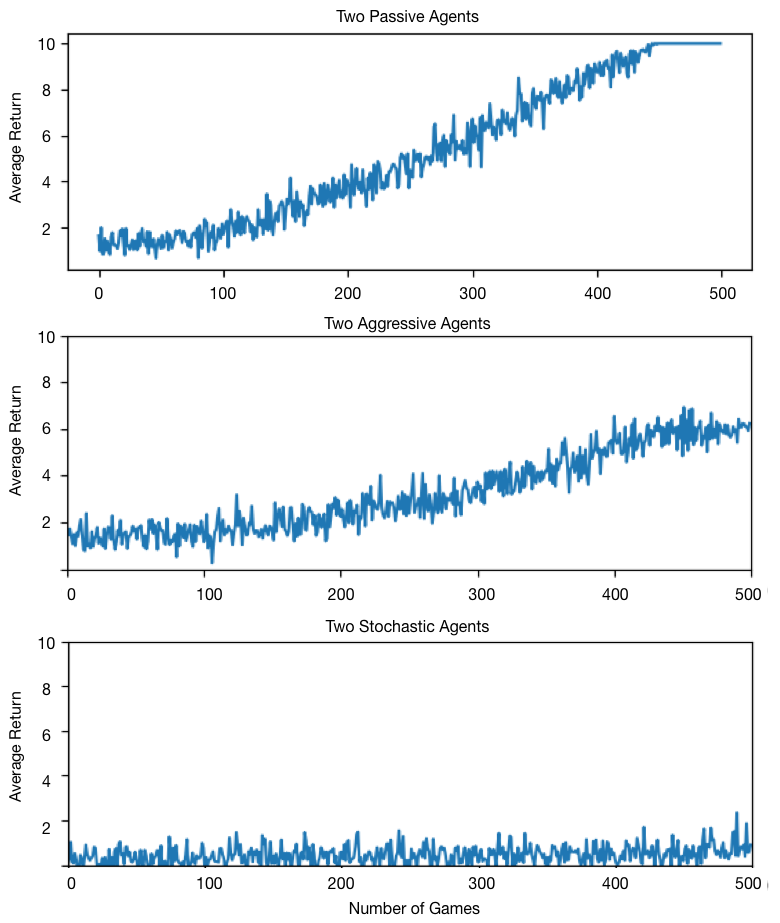}
\caption{Learning curves for the Q-learning using the behavior model with $\alpha=0.001$ and $\gamma=0.98$.  The optimal policy is learned for the passive agents, and a reasonable policy is learned for the aggressive agents.  However, a policy cannot be learned for multiple stochastic agents.}
\label{fig:MultipleAgentsSameType}
\end{figure}

Figure \ref{fig:MultipleAgentsSameType} (top) shows that if there are multiple agents of the same type that behave well, in terms of turn-taking interactions and honor the intent signaled by other agents, then the Q-learning agent is able to learn an optimal policy and take turns appropriately. Furthermore, the rate at which the learning occurs is the same as if there is a single passive agent present (c.f.\ the top of Figure \ref{fig:PassiveAgentLearningCurve}).

As the scenario becomes more challenging, the Q-learning agent no longer learns an optimal policy.  For the case of multiple aggressive agents the likelihood of collisions increases since they always take block placing actions after indicating, regardless of the state of other agents.  Despite this behavior the Q-learning agent is able to learn a reasonable policy as the weights on the output arcs of the WFST converge to the distributions defining the behavior of the agents (middle of Figure \ref{fig:MultipleAgentsSameType}).  As one might expect, for the complex case of multiple stochastic agents (bottom of Figure \ref{fig:MultipleAgentsSameType}) the Q-learning agent is not able to learn a policy.  This limitation is because it becomes increasingly difficult to predict how the world will evolve as it becomes increasingly stochastic.

\section{Conclusions and Future Work}\label{sec:conclusions}

We have presented a model of turn-taking behavior for individual and groups of agents, and we have shown that a learning-agent can use the model to select turn-taking actions. The hierarchical nature of our behavior model and our treatment of all rule-based agents as a single group unit allows us to learn a turn-taking policy for interacting with an \emph{arbitrary numbers of agents}.  Our model is dependent only on the number of behavior \emph{types} that the learning-agent is likely to encounter.  The accuracy of the action selection policy is dependent on the stochasticity of the agents.  However, with a reasonable degree of regularity to the agents (as we would expect in real-world tasks), the policy is able to successfully and safely take turns.

Future work could consider continuous domains and handling time-pressure conditions by replacing the binary intention signal with a measure of action completion and the speed of action execution. The measure could operate as an abstraction similar to that used by \citep{raux:2007:arch}.  We could also replace the Q-learners used in this work with a learning algorithm that has been shown to work in continuous domains, e.g.\ deep deterministic policy gradient for robotics control \citep{popov:2017:robot}.

In this work we used rule-based agents where each agent was represented by a single and known \emph{type}. When playing against either humans or agents that we did not design, we do not know \emph{a priori} their behavior types. Therefore, we would need to extend our use of WFSTs by generating hypotheses about agent \emph{type} and update these based on how well each model of behavior \emph{type} is able to predict the behaviors of the other agents.  Alternatively, we might model an agent as a distribution over agent \emph{types} rather than imposing a strict classification.  Furthermore, agents could be assigned initially to an ``unknown'' behavior \emph{type}, and learning-agents take turns with caution.  As the behavior models are refined during interactions, the learning-agents can be more confident in their predictions.

In summary, the performance of our model in the tower-building domain shows it is encouraging for its application to learning to interact with an arbitrary numbers of agents of different behavioral \emph{types} in collaborative environments.

\section{Acknowledgments}
The authors are grateful to Russ Webb, Yin Zhou, Megan Maher and Melanie Subbiah for their helpful discussions and valuable comments on the work.

\bibliographystyle{plain}  
\bibliography{bibliography} 

\end{document}